\documentclass[journal=jacsat,manuscript=article]{achemso}

\usepackage[version=3]{mhchem} % Formula subscripts using \ce{}

\usepackage[caption=false]{subfig}
\usepackage{pgfplots, pgfplotstable}
\usepackage[colorlinks = true,
            linkcolor = blue,
            urlcolor  = blue,
            citecolor = blue,
            anchorcolor = blue]{hyperref}
\pgfplotsset{compat=1.17}
\author{Youjun Xu}
\affiliation[iipharma]
{Infinite Intelligence Pharma, Beijing, China, 100083}
\altaffiliation{These authors contributed equally}
\email{xuyj@iipharma.cn}

\author{Jinchuan Xiao}
\affiliation[iipharma]
{Infinite Intelligence Pharma, Beijing, China, 100083}
\altaffiliation{These authors contributed equally}

\author{Chia-Han Chou}
\affiliation[iipharma]
{Infinite Intelligence Pharma, Beijing, China, 100083}

\author{Jianhang Zhang}
\affiliation[iipharma]
{Infinite Intelligence Pharma, Beijing, China, 100083}

\author{Jintao Zhu}
\affiliation[cqb]
{Center for Quantitative Biology, Peking University, Beijing, China, 100871}

\author{Qiwan Hu}
\affiliation[cqb]
{Center for Quantitative Biology, Peking University, Beijing, China, 100871}

\author{Hemin Li}
\affiliation[iipharma]
{Infinite Intelligence Pharma, Beijing, China, 100083}

\author{Ningsheng Han}
\affiliation[iipharma]
{Infinite Intelligence Pharma, Beijing, China, 100083}

\author{Bingyu Liu}
\affiliation[iipharma]
{Infinite Intelligence Pharma, Beijing, China, 100083}

\author{Shuaipeng Zhang}
\affiliation[iipharma]
{Infinite Intelligence Pharma, Beijing, China, 100083}

\author{Jinyu Han}
\affiliation[iipharma]
{Infinite Intelligence Pharma, Beijing, China, 100083}

\author{Zhen Zhang}
\affiliation[iipharma]
{Infinite Intelligence Pharma, Beijing, China, 100083}

\author{Shuhao Zhang}
\affiliation[iipharma]
{Infinite Intelligence Pharma, Beijing, China, 100083}

\author{Weilin Zhang}
\affiliation[iipharma]
{Infinite Intelligence Pharma, Beijing, China, 100083}

\author{Luhua Lai}
\fax{(+86)10-62751725}
\affiliation[cqb]
{Center for Quantitative Biology, Peking University, Beijing, China, 100871}
\alsoaffiliation[ccme]
{BNLMS, Peking-Tsinghua Center for Life Sciences at the College of Chemistry and Molecular Engineering, Peking University, Beijing, China, 100871}
\email{lhlai@pku.edu.cn}

\author{Jianfeng Pei}
\fax{(+86)10-62759595}
\affiliation[cqb]
{Center for Quantitative Biology, Peking University, Beijing, China, 100871}
\email{jfpei@pku.edu.cn}

\title{MolMiner: You only look once for chemical structure recognition}

\abbreviations{YOLO: You look only once; OCSR: Optical chemical structure recognition; OCR: Optical character recognition; CLiDE: Chemical literature data extraction;  }

\begin{document}

% \begin{tocentry}

% Some journals require a graphical entry for the Table of Contents.
% This should be laid out ``print ready'' so that the sizing of the

% Inside the \texttt{tocentry} environment, the font used is Helvetica
% 8\,pt, as required by \emph{Journal of the American Chemical

% The surrounding frame is 9\,cm by 3.5\,cm, which is the maximum
% permitted for  \emph{Journal of the American Chemical Society}
% graphical table of content entries. The box will not resize if the
% content is too big: instead it will overflow the edge of the box.

% This box and the associated title will always be printed on a
% separate page at the end of the document.

% \end{tocentry}

\begin{abstract}
Molecular structures are always depicted as 2D printed form in scientific documents like journal papers and patents. However, these 2D depictions are not machine-readable. Due to a backlog of decades and an increasing amount of these printed literature, there is a high demand for the translation of printed depictions into machine-readable formats, which is known as Optical Chemical Structure Recognition (OCSR). Most OCSR systems developed over the last three decades follow a rule-based approach where the key step of vectorization of the depiction is based on the interpretation of vectors and nodes as bonds and atoms. Here, we present a practical software MolMiner, which is primarily built up using  deep neural networks originally developed for semantic segmentation and object detection to recognize atom and bond elements from documents. These recognized elements can be easily connected as a molecular graph with distance-based construction algorithm. We carefully evaluate our software on four benchmark datasets with the state-of-the-art performance. Various real application scenarios are also tested, yielding satisfactory outcomes. The free download links of Mac and Windows versions are available:   \href{https://molminer-cdn.iipharma.cn/pharma-mind/artifact/latest/mac/PharmaMind-mac-latest-setup.dmg}{Mac link} and \href{https://molminer-cdn.iipharma.cn/pharma-mind/artifact/latest/win/PharmaMind-win-latest-setup.exe}{Windows link}.
\end{abstract}

\section{Introduction}
Chemical products are a vast amount of priceless wealth, and are making our life and health better. Much efforts on chemical research and development have been made and published as primary scientific literature. During the last decade, researchers have developed various machine learning and deep learning models to solve a series of predictive and generative tasks in the fields of chemistry and biology.\cite{xu2019deep,lavecchia2019deep,tang2019recent} It is obvious that well-performed computational models cannot be separated from data accumulation, especially for experimental data, \textit{e.g.} chemical reaction data and biological active data. 

Some well-known databases have play a vital role on scientific studies. For example, the Protein Data Bank database is an important undertaking to make protein crystal structural data to be public, which has greatly facilitated research efforts and knowledge developments.\cite{burley2021rcsb} Several large comprehensive biomedical databases like ChEMBL have been constructed and updated. These datasets offer a necessary basis and chance to develop various practical advanced technologies.\cite{mendez2019chembl} Recently, there is a renewed interest to structurally collate experimental data sets and build their inter- and intra- relationships to enhance various downstream predictions and recommendations. The Open Reaction Database aims to collect and share chemical reaction data from journal articles, patents, and even electronic laboratory notebooks.\cite{kearnes2021open} Due to the rapidly increasing amount of literature resources, we find it both laborious and time-consuming to integrate diverse kinds of experimental data into a comprehensive and professional knowledge database.

Automatic computational methods provide a potential option to handle with various forms of valuable chemical and biological information. Named Entity Recognition tools have been applied to extract chemical textual information from the literature to create structured data.\cite{eltyeb2014chemical} Besides text-like objects, researchers also have developed Optical Chemical Structure Recognition (OCSR) tools with the intention of decoding a graphical chemical depiction into a machine-readable molecular format. However, there is still a non-trivial problem of how to accurately recognize the molecular structures from electronic materials.

Since 1990, several closed- and open-source OCSR systems have been established based on similar rule-based implementations involving image vectorization, image thinning, line enhancement, text-based Optical Character Recognition (OCR) and graph reconstruction. A representative system, named Chemical Literature Data Extraction (CLiDE), is a commercial OCSR toolkit developed by Keymodule.\cite{valko2009clide} And it has been integrated into ChemAxon software.\cite{chemaxon} Generally, most of the commercial OCSR systems were unavailable to academic researchers. In 2009, Filippov and Nicklaus published the first open-source system called Optical Structure Recognition Application (OSRA) \cite{filippov2009optical}. And this tool is kept active and upgraded for improved recognition. Imago and MolVec also have been developed as open-source systems to offer researchers optional tools of molecular structure recognition.\cite{imago,molvec} More detail OCSR systems can refer to this review\cite{ocsrreview}. 

Dramatic developments on deep learning (DL) frameworks and hardware have achieved in image recognition technologies. Recently, a biopharmaceutical company, Bristol-Myers Squibb, held a competition of molecular translation in Kaggle.\cite{bmsmolecular} The architecture of CNN-Transformer plays an essential role in translating chemical images to InChI strings\cite{heller2013inchi}. Based on this, DECIMER has been reported for translating various chemical images to SELFIES strings with an acceptable precision.\cite{rajan2021decimer} Similarly, Bayer researchers have developed another translation method called Img2Mol, exhibiting a potential ability of recognizing hand-drawn molecules.\cite{clevert2021img2mol} Inspired by image segmentation technologies, ChemGrapher used atom-based, bond-based and charge-based segmentation neural networks to predict the probabilities of each pixel for a chemical image, and then construct a chemical graph.\cite{oldenhof2020chemgrapher} Following this work, ABC-Net applied a divide-and-conquer segmentation strategy to significantly improve recognition performance.\cite{zhang2022abc} Although these DL-based works deliver a promising and potential application value, these results actually are short of rigorous evaluations on benchmark datasets and real-world data sets.

Taking substantial strengths of DL into consideration\cite{lecun2015deep}, we developed a practical DL-based system called MolMiner to improve OCSR tasks. It can rapidly and accurately extract chemical images and then recognize chemical structures from PDF-format documents. MolMiner was also fairly tested using the benchmark data sets\cite{ocsrreview} and some real-world datasets. The test performance shows that MolMiner can outperform three existing open-source systems. We also integrated some basic functions like real-time correction function and screenshot function into MolMiner to provide a user-friendly interface. The Win/Mac free download links are \href{https://molminer-cdn.iipharma.cn/pharma-mind/artifact/latest/win/PharmaMind-win-latest-setup.exe}{Windows link} and  \href{https://molminer-cdn.iipharma.cn/pharma-mind/artifact/latest/mac/PharmaMind-mac-latest-setup.dmg}{Mac link}, respectively. MolMiner would be freely available with daily permission to all registered users.

\section{MolMiner Recognition System}
\subsection{Implementation}
MolMiner is a rule-free learning system. It aims to transform the vectorization problem into object detection tasks. That means it can extract chemical elements in a object detection manner by training well-labelled datasets with atom and bond annotations. It was implemented by five main modules as follows.

\begin{itemize}
\item Data generation and annotation module aims to automatically generate various styles of well-annotated chemical images based on the modified RDKit toolkit.\cite{rdkiturl} It can support several augmentation operations, such as rotation, thinning, thickness, noise and super group. More augmentation operations (e.g. hand-drawn lines) will be added into this module in the future, which is well-suited for various real application scenarios. 

\item Chemical image extraction module is a fully DL-driven image segmentation module, and is implemented by a light-weight model MobileNetV2.\cite{sandler2018mobilenetv2} It is used to extract printed chemical depiction from PDF-format documents. Labelled data is generated by the first module with several predefined templates such as journal style and patent style. The two categorical annotations are adopted to train this model with the weighted cross entropy loss of a ratio 1:2 of background-class and compound-class. The key performace of recall is 95.5\%.  

\item Chemical image recognition module is the paramount key module for rapid and accurate chemical structure recognition. It is implemented based on a popular one-stage YOLOv5 architecture\cite{glennjocher}. After a series of evaluations on MaskRCNN\cite{he2017mask}, FastRCNN\cite{girshick2015fast}, EfficientDet\cite{tan2020efficientdet}, we empirically found the YOLOv5 model could significantly outperform other object detection architectures just for this recognition task. The atom labels include  ``Si", ``N", ``Br", ``S", ``I", ``Cl", ``H", ``P", ``O", ``C", ``B", ``F", ``Text". And the bond labels include ``Single", ``Double", ``Triple", ``Wedge", ``Dash", ``Wavy". The whole predicted mAP@.5 is 97.5\%. Labelled data is automatically generated by the first module with several image-level augmentation operations.

\item Text-based optical character recognition (OCR) module is used to recognize chemistry-text images with atom characters and super groups. This OCR model is implemented by finetuning a pretrained EasyOCR model\cite{easyocr} with specifically cropped chemical texts from the first module. The accuracy performance is about 96.4\%.

\item Chemical element construction and evaluation module contains a distance-based graph construction algorithm accompanied by super group parser, and an automatic evaluation module for performing fair MCS-based and InChI-based comparisons on benchmark datasets. The super groups are collected from RDKit code, ChemAxon, OSRA and some JMC journals. The evaluation results are discussed in the Section of Benchmark Evaluation.
\end{itemize}

In comparison to rule-based systems, MolMiner has several advantages: i) Batch GPU-based inference can make the speed of element recognition faster than vectorization algorithms; ii) It can implicitly learn rules from a large amount of automatic generated datasets or other manual annotated datasets, without designing explicit rules. This feature can be simply extended to new scenes; iii) It can support rapid synchronous recognition of multiple large-size images; iv) It provides additional user-friendly interfaces for easy-to-use.

\subsection{User Interface}
User interface of MolMIner integrate several frequently-used functions including screenshot recognition, batch PDF recognition, real-time molecular edit, collection, and SDF/XLSX-format download. 

\begin{figure}[H]
\centering 
\includegraphics[width=\textwidth]{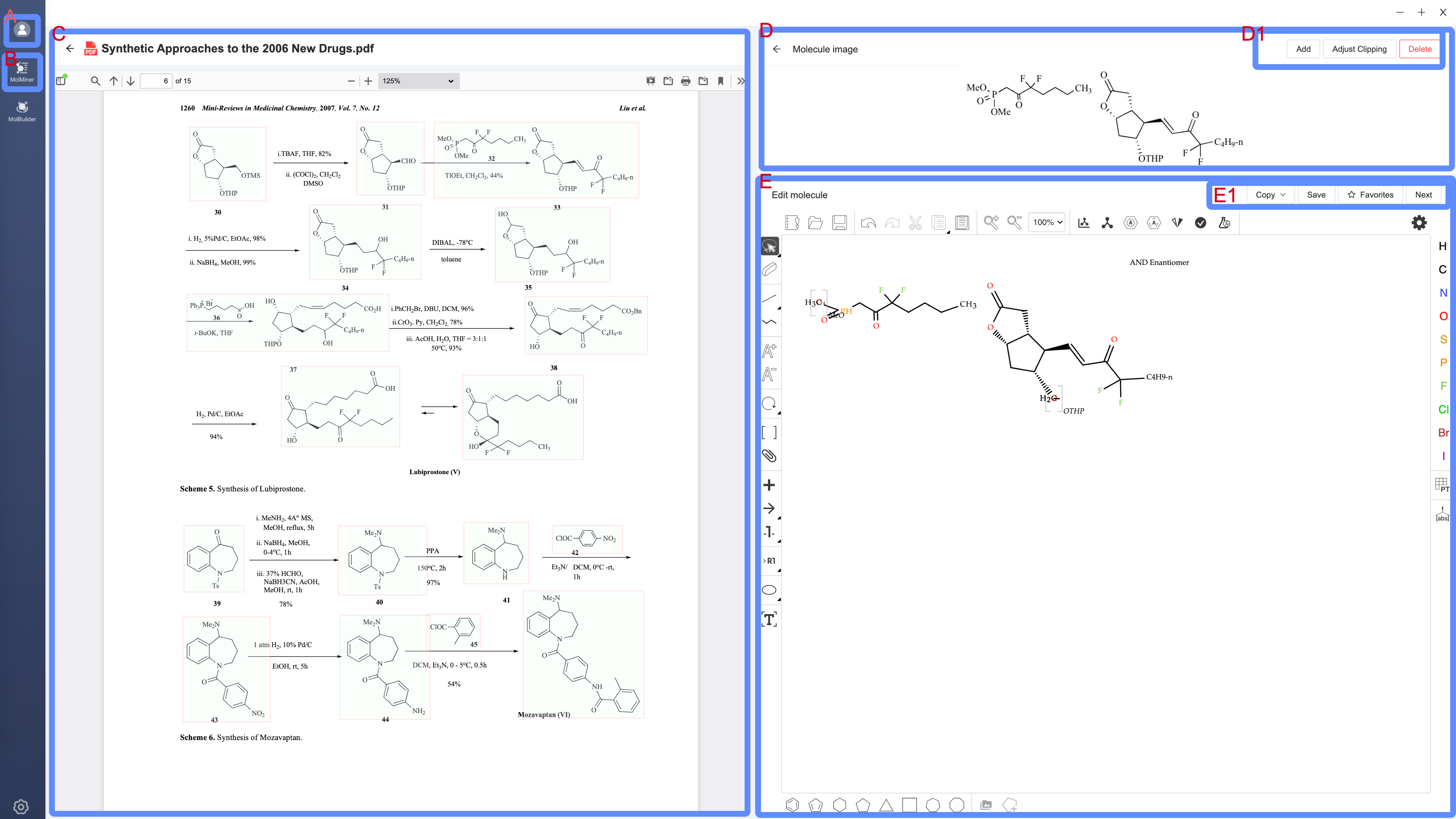}
\caption{Edit interface with five rectangle parts (in blue). A) User's account information; B) MolMiner module; C) PDF viewer (supported by PDF.js v2.14\cite{pdfviewer}); D) Cropped molecular image. D1 contains three buttons. The ``Add" function is to add another cropped image of interest. The ``Adjust Cropping" is to adjust the boundaries of the cropped images. The ``Delete" is to remove the selected cropped image; E) Real-time molecular edit (supported by Ketcher v2.4\cite{ketcher}). E1 contains the four basic functions of Copy, Save, Favorites and Next.}
\label{fig:UIEditor}
\end{figure}

Figure \ref{fig:UIEditor} presents the overview of real-time molecular edit interface. In this interface, user can check and correct recognized molecules one-by-one to achieve the perfect accuracy of 100\%. After a simple testing on 1000 images of non-markush organic medicinal molecules, it took about 5-10s for one simply-trained person to process one molecule with check, correct and save operations. Other functional modules can be easily found in the MolMiner software. Details can refer to the MolMiner User Manuals.

\section{MolMiner Evaluation}
\subsection{Benchmark Evaluation} 
We tested four benchmark data sets (USPTO, UOB, JPO, CLEF2012) referred to \citet{ocsrreview}, shown in Table \ref{tab:datasets}. We compared MolMiner's runtime and accuracy performance with other existing open-source OCSR tools such as MolVec\cite{molvec}, OSRA\cite{filippov2009optical} and Imago\cite{imago} on these four datasets. We reviewed the consistency of InChI strings that is mentioned in \citet{ocsrreview}. Considering the atom-level evaluation method from Imago,\cite{imagoreport} we found that the atom-level and bond-level consistency index of maximum common structures (MCS) should be an appropriate metric for evaluating this task. It can comprehensively measure the accuracy of atom-level and bond-level. In Table \ref{tab:LSDvsMCS}, we compared InChI-based accuracy and MCS-based accuracy, and found some significant differences due to the count of None value, bond misplacement in aromatic rings, and RDKit-based InChI export. Here, MCS-based accuracy is deemed as a more rigorous indicator to monitor the fine-grained accuracy of atoms and bonds. Successively, we analyzed the runtime and MCS-based (both atom-level and bond-level) accuracy between our recognized molecules and ground truth molecules. The results are summarized in Table \ref{tab:LSDvsMCS}, showing that MolMiner can outperform the open-source tools both in runtimes and MCS accuracies on USPTO, UOB, JPO and CLEF2012.  

Based on rule-based strategies, CLiDE Pro is a popular and professional commercial software. Unfortunately, we currently cannot perform a fair comparison on this software. According to reported USPTO results from ChemAxon,\cite{keymodulereport} the reported accuracy (93.8\%) is slightly better than our MCS-based accuracy (93.3\%). One of possible reasons is the issue of crossing bonds. As previously mentioned, since MolMiner is based on DL-based object detection models, it is difficult for crossing bonds to be well recognized just by these simple atom and bond annotations. Manual correction is recommended in the interactive plugin of Ketcher. We are planning to solve this issue with several unavoidable rules or coarse-grained annotations.

\begin{table}[htbp]
\centering
\renewcommand{\arraystretch}{1.3}
\begin{tabular}{cccccccc}
Dataset &  & Size  & &  MW avg.   &  & MW std. \\ \hline\hline 

USPTO    &  &  5719 & &  379.95 &  &  115.38  \\ \hline

UOB      &  &  5740 & &  213.49 &  &  57.30 \\ \hline

JPO      &  &  450  & &  360.27 &  &  184.77 \\ \hline

CLEF2012 &  &  992  & &  401.16 &  &  143.89 \\ \hline
\end{tabular}
\caption{A summary of benchmark dataset sizes and molecular weights (MW)}
\label{tab:datasets}
\end{table}

\begin{table}[htbp]
\centering

\renewcommand{\arraystretch}{1.3}
\begin{tabular}{ccccccccc}
Dataset & & Molvec& & Imago &  & Osra & &  MolMiner\\ \hline\hline 

USPTO & \begin{tabular}[c]{@{}c@{}} InChI acc. (\%)\\ MCS acc. (\%) \end{tabular} & \begin{tabular}[c]{@{}c@{}} 89 \\ 89\end{tabular} & &  \begin{tabular}[c]{@{}c@{}} 86 \\ 86 \end{tabular} & &  \begin{tabular}[c]{@{}c@{}} 88 \\ 17 \end{tabular} & & \begin{tabular}[c]{@{}c@{}} \textbf{90} \\ \textbf{93} \end{tabular} \\ \hline

UOB  & \begin{tabular}[c]{@{}c@{}} InChI acc. (\%) \\ MCS acc. (\%) \end{tabular} & \begin{tabular}[c]{@{}c@{}} 88 \\ 49\end{tabular} & &  \begin{tabular}[c]{@{}c@{}} 48\\ 36 \end{tabular} & &  \begin{tabular}[c]{@{}c@{}} 87 \\ 27 \end{tabular}   & & \begin{tabular}[c]{@{}c@{}} \textbf{90}\\ \textbf{63} \end{tabular} \\ \hline

CLEF2012 & \begin{tabular}[c]{@{}c@{}} InChI acc. (\%) \\ MCS acc.(\%) \end{tabular} & \begin{tabular}[c]{@{}c@{}} 84 \\ 83 \end{tabular} & &  \begin{tabular}[c]{@{}c@{}} 58 \\ 60 \end{tabular} & &  \begin{tabular}[c]{@{}c@{}} 84 \\ 22 \end{tabular}  & & \begin{tabular}[c]{@{}c@{}} \textbf{85} \\ \textbf{87} \end{tabular} \\ \hline

JPO & \begin{tabular}[c]{@{}c@{}} InChI acc. (\%) \\ MCS acc. (\%) \end{tabular} & \begin{tabular}[c]{@{}c@{}} 66 \\ 32 \end{tabular} & &  \begin{tabular}[c]{@{}c@{}} 40 \\ 26 \end{tabular} & &  \begin{tabular}[c]{@{}c@{}} 66 \\ 20 \end{tabular}   & & \begin{tabular}[c]{@{}c@{}} \textbf{72}\\ \textbf{35} \end{tabular}\\ \hline

\end{tabular}
\caption{An overview of InChI-based and MCS-based accuracy performance}
\label{tab:LSDvsMCS}
\end{table}

\begin{table}[htbp]
\centering
\renewcommand{\arraystretch}{1.3}
\begin{tabular}{cccccccc}
Datasets & & Molvec  & Imago   & Osra  & MolMiner \\ \hline\hline 

USPTO & \begin{tabular}[c]{@{}c@{}} Runtime (min) $\downarrow$\\ MCS acc. (\%) $\uparrow$\end{tabular} & \begin{tabular}[c]{@{}c@{}} 29  \\ 89 \end{tabular}  &  \begin{tabular}[c]{@{}c@{}} 73  \\ 86 \end{tabular}  &  \begin{tabular}[c]{@{}c@{}} 148 \\ 17 \end{tabular}  &\begin{tabular}[c]{@{}c@{}} \textbf{7} \\ \textbf{93}\end{tabular} \\ \hline

UOB  & \begin{tabular}[c]{@{}c@{}}  Runtime (min) $\downarrow$\\ MCS acc. (\%) $\uparrow$\end{tabular} & \begin{tabular}[c]{@{}c@{}} 28  \\ 49 \end{tabular}  &  \begin{tabular}[c]{@{}c@{}} 153  \\ 36 \end{tabular}  &  \begin{tabular}[c]{@{}c@{}} 126  \\ 27 \end{tabular}  & \begin{tabular}[c]{@{}c@{}} \textbf{6} \\ \textbf{63}\end{tabular} \\ \hline

CLEF2012 & \begin{tabular}[c]{@{}c@{}} Runtime (min) $\downarrow$\\ MCS acc. (\%) $\uparrow$\end{tabular} & \begin{tabular}[c]{@{}c@{}} 4  \\ 83 \end{tabular}  &  \begin{tabular}[c]{@{}c@{}} 16  \\ 60 \end{tabular}  &  \begin{tabular}[c]{@{}c@{}} 21 \\ 22 \end{tabular}  &\begin{tabular}[c]{@{}c@{}} \textbf{1} \\ \textbf{87} \end{tabular} \\ \hline

JPO  & \begin{tabular}[c]{@{}c@{}}  Runtime (min) $\downarrow$\\ MCS acc. (\%) $\uparrow$\end{tabular} & \begin{tabular}[c]{@{}c@{}} 8 \\ 32 \end{tabular}  &  \begin{tabular}[c]{@{}c@{}} 23 \\ 26\end{tabular}  &  \begin{tabular}[c]{@{}c@{}} 17  \\ 20 \end{tabular}  &\begin{tabular}[c]{@{}c@{}} \textbf{$<$1} \\ \textbf{35}\end{tabular} \\ \hline

\end{tabular}
\caption{A summary of benchmark evaluation on runtime and MCS accuracy}
\label{tab:benchmarks}
\end{table}

\subsection{Application Case Evaluation}

Currently, MolMiner focuses on processing organic medicinal molecules (Non-Markush structures) from scientific documents. Molecular images with some errors (including crossing bonds, noise and super group parser) are still needed some manual corrections in the interactive plugin of Ketcher. Here, we proposed and tested three application cases to validate the ability of MolMiner.
\begin{itemize}
\item We took screenshot of a large-size image (3000 $\times$ 2068, 300 dpi)  of palytoxin to make an test. It took approximately 3 seconds to return one well-recognized molecular structure without any manual corrections in the Ketcher plugin (shown in Figure \ref{fig:largesizecase}). 

\begin{figure}[htbp]
\centering 
\includegraphics[width=\textwidth]{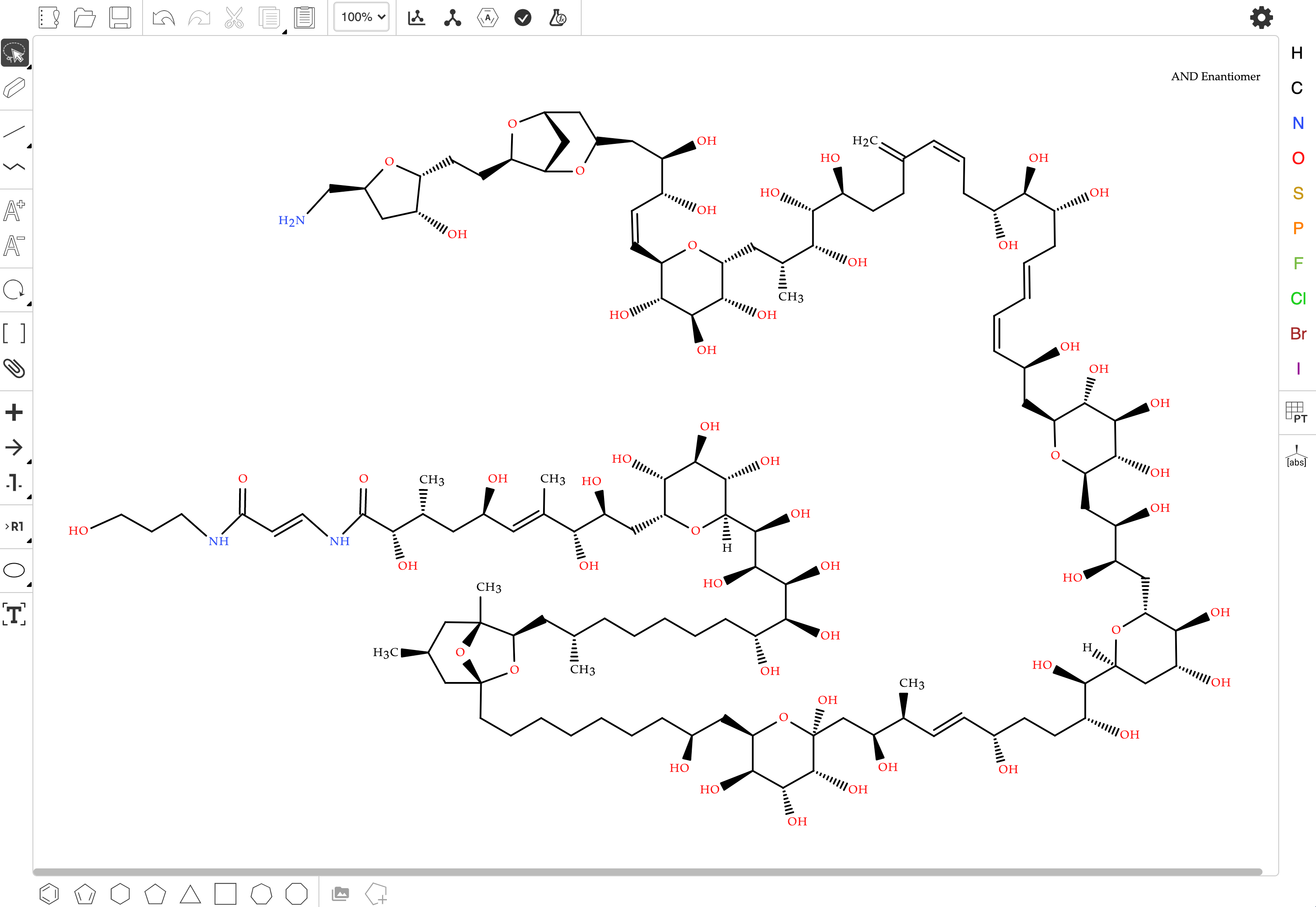}
\caption{A large-size (3000 $\times$ 2068) case of palytoxin}
\label{fig:largesizecase}
\end{figure}

\item A scanned PDF page from a scientific journal\cite{wang2021traditional} (machine: HP M1210 MFP, dpi: 300, brightness: 128, contrast: 124) was tested by MolMiner, shown in Figure \ref{fig:pdfcase}. The recognized results are satisfactory and achieve 100\% accuracy under a set of appropriate scanning parameters. We also tried some sets of scanning parameters, which more or less could influence the recognized results.

\begin{figure}[!htbp]
\centering 
\includegraphics[width=0.85\textwidth]{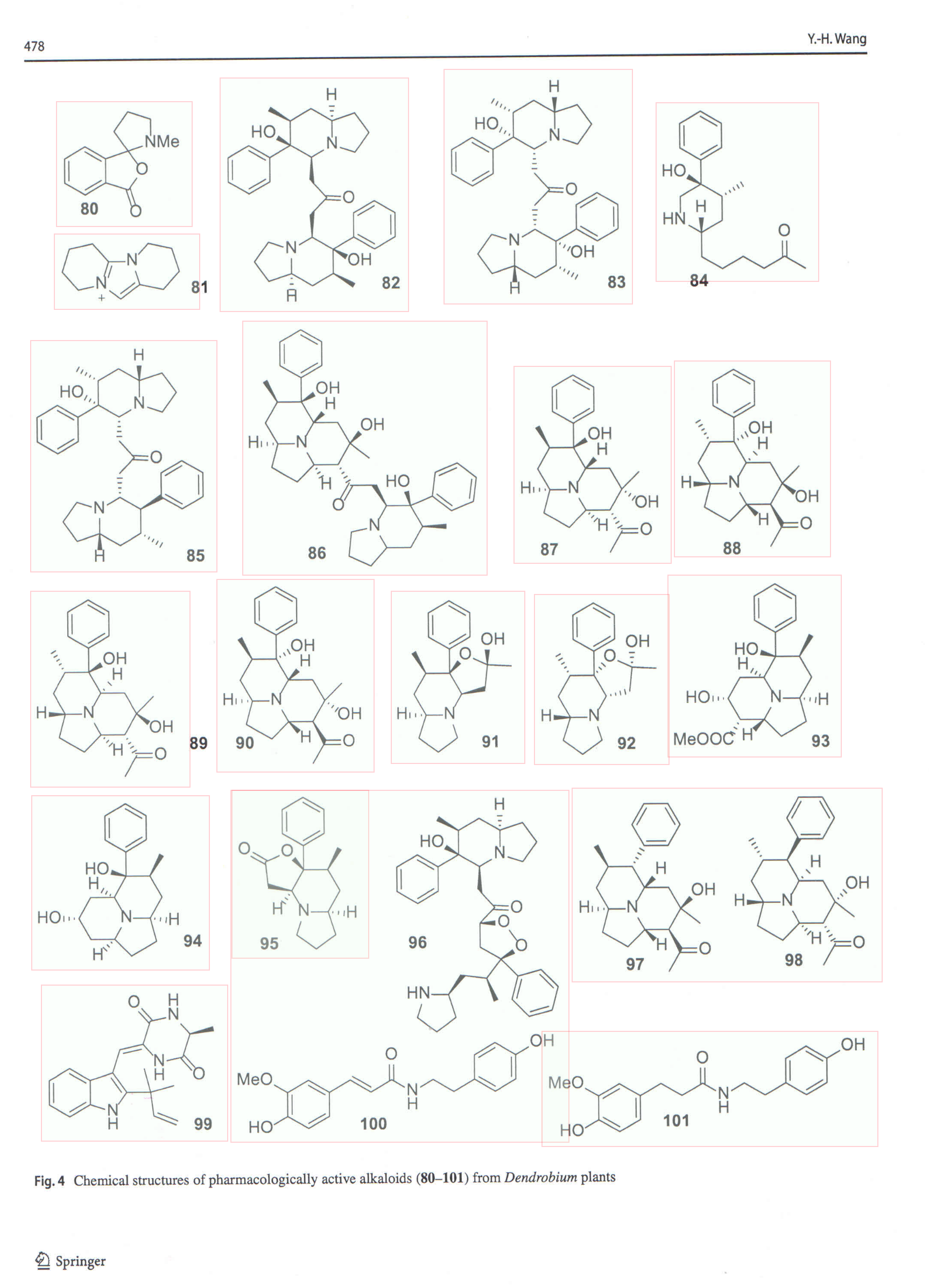}
\caption{A case of one scanned journal page from \citet{wang2021traditional}}
\label{fig:pdfcase}
\end{figure}

\item We tested a challenging case of hand-drawn images from \citet{clevert2021img2mol}. For Figure \ref{fig:handcase}, although there are still some errors including ``N", ``Cl", wedge bonds and aromatic rings, MolMiner (without training any similar data) could afford to recognize some simple hand-drawn images (Figure \ref{fig:handcase}(right)), and main skeletons and their positions of some complex images (Figure \ref{fig:handcase}(left)). As a matter of fact, the retained position layout is beneficial for users to check and correct their structures. 

\begin{figure}[htbp]
\centering 
\includegraphics[width=\textwidth]{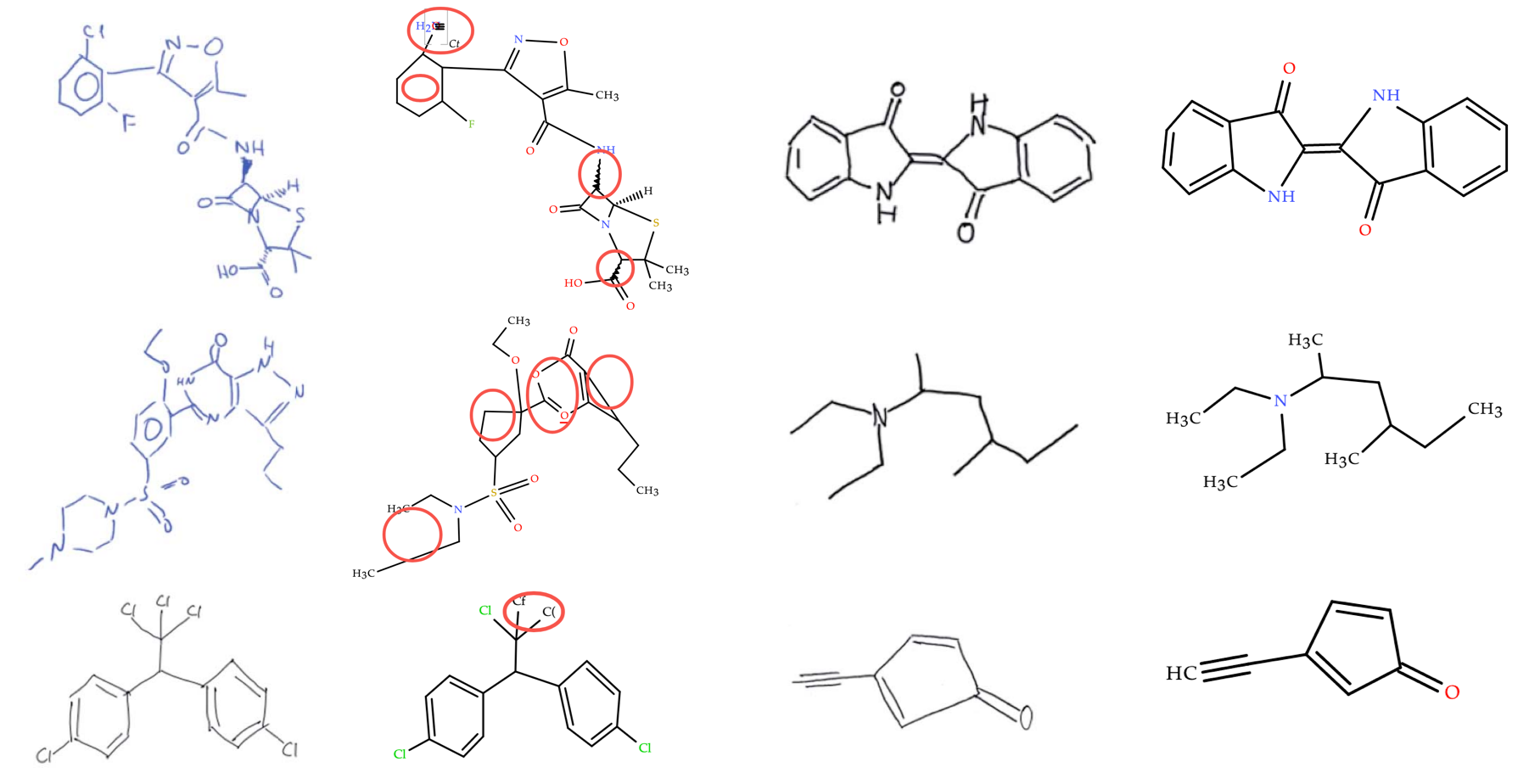}
\caption{A case of hand-drawn images from \citet{clevert2021img2mol}. Some errors are highlighted in red, including ``N", ``Cl", wedge bonds and aromatic rings}
\label{fig:handcase}
\end{figure}
\end{itemize}

\subsection{Limitations}
Current version of MolMiner is weak at dealing with the problems including crossing bonds, colorful backgrounds, crowded layout segmentation and Markush structures. These problems still need more or less manual corrections in the Ketcher plugin.

\section{Conclusion}
We have developed a practical OCSR software called MolMiner based on advanced deep learning technologies (such as MobileNetv2, YOLOv5, EasyOCR). We also developed an automatic data generation module to satisfy the data volume requirements of DL models. The benckmark evaluation suggested current MolMiner could outperform the open-source OCSR tools (\textit{e.g.} MolVec, OSRA, Imago) on the performance of both accuracy and runtime. MolMiner supports several frequently-used functions of batch PDF recognition, screenshot recognition and real-time molecular edit. More other functions will be gradually integrated into MolMiner. MolMiner is freely available here\cite{downloadlink} with Mac and Windows versions. We open daily access without cost for all registered users to ensure daily usage amount. We also provide extra channels of enterprises for unlimited access.

\begin{acknowledgement}
This work is supported by the funding from Infinite Intelligence Pharma Ltd.
\end{acknowledgement}


\bibliography{achemso-demo}
\end{document}